\documentclass[11pt,a4paper]{article}
\usepackage[hyperref]{emnlp-ijcnlp-2019}
\usepackage{times}
\usepackage{latexsym}

\usepackage{url}
\usepackage{threeparttable}
\usepackage{tabularx}
\usepackage{graphicx}
\usepackage{float}
\usepackage{xspace}
\usepackage{footmisc}
\usepackage{booktabs}
\usepackage{tablefootnote}
\usepackage{amsmath}

\aclfinalcopy 

\newcommand{\SentApp}{ABSApp\xspace}

\title{\SentApp: A Portable Weakly-Supervised \\ Aspect-Based Sentiment Extraction System}

\author{Oren Pereg$^1$, Daniel Korat$^1$, Moshe Wasserblat$^1$, Jonathan Mamou$^1$, Ido Dagan$^2$\\
  $^1$Intel AI Lab, Petah Tikva, Israel \\
  $^2$Department of Computer Science, Bar-Ilan University,  Ramat Gan, Israel \\
  $^1${\tt firstname.lastname@intel.com} \\
  $^2${\tt dagan@cs.biu.ac.il}\\
}

\date{}

\begin{document}
\maketitle
\begin{abstract}
We present \SentApp, a portable system for weakly-supervised aspect-based sentiment extraction\footnote{A demo video of \SentApp is available at \url{https://drive.google.com/open?id=1BLk0xkjIOqyRhNy4UQEFQpDF_KR_NMAd}.}. The system is interpretable and user friendly and does not require labeled training data, hence can be rapidly and cost-effectively used across different domains in applied setups. The system flow includes three stages: First, it generates domain-specific aspect and opinion lexicons based on an unlabeled dataset; second, it enables the user to view and edit those lexicons (weak supervision); and finally, it enables the user to select an unlabeled target dataset from the same domain, classify it, and generate an aspect-based sentiment report. \SentApp has been successfully used in a number of real-life use cases, among them movie review analysis and convention impact analysis.
\end{abstract}

\section{Introduction}
\label{sec:introduction}

\begin{figure*}[ht]
\includegraphics[width=\textwidth]{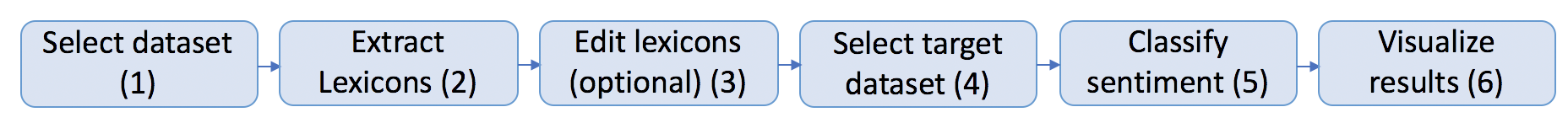}
\centering
\caption{\SentApp workflow.}
\label{workflow_fig}
\end{figure*}

Aspect Based Sentiment Analysis (ABSA) is the task of extracting, from a given corpus, opinion targets (aspect terms) and the sentiment expressed towards them. For example, in the sentence {\it``The dessert was incredible''}, the aspect term is {\it dessert} and the sentiment towards it is positive. 
This fine-grained trait of ABSA makes it an effective application for measuring and monitoring the ratio between positive and negative opinions expressed towards specific aspects of a product or service.

Most work around ABSA focused on supervised sequence tagging based systems. \citet{Liu2015in_domain_co_extraction}
showed promising results when the training and the inference data are from the same domain. However, this approach is typically not robust across different domains since aspect terms from two different domains are usually semantically different hence separated in the embedding vector space. For example, frequent aspect terms in the restaurant domain, like {\it food, menu, starters} and {\it salad}, have little or no semantic relatedness to frequent aspect terms in the laptop domain, like {\it screen size, keyboard} and {\it battery life}. In addition to aspect terms, many opinion terms are also domain-specific. For example, opinion terms like {\it tasty (positive), yummy (positive), flavorful (positive)} and {\it tasteless (negative)} are specific to the restaurant domain whereas opinion terms like {\it lightweight (positive), durable (positive), compatible (positive)} and {\it heavy (negative)} are specific to the laptop domain.
This makes domain-agnostic ABSA a challenging task, with little work addressing it. 

A recent line of work is based on transfer-learning methods, in which labeled data from a source domain is used for training a model to classify data in a target domain. \citet{Ding2017crossdomain} and \citet{Wang2018cross_domain} proposed using supervised RNNs for cross-domain aspect term extraction and for aspect and opinion term co-extraction. This approach showed encouraging results, however it requires a considerable amount of labeled data from the source domain which is often not practical in applied settings due to cost or legal considerations (relevant data is usually not available for commercial use). 

Another approach towards domain robustness is based on unsupervised methods. \citet{Hu_and_Lui2004rule_based} used association rules and \citet{Qiu_and_Lui2011double_prop} used syntactic rules for aspect and opinion term co-extraction. 
Industrial setups usually lack labeled data for training and this is where unsupervised methods excel. However, these methods can be noisy (see the \SentApp-unsup baseline in Table~\ref{table:ATE_table}). In this paper we show that weak supervision, namely a short manual process of editing lexicons that were generated by an unsupervised method, produces results that are comparable to transfer-learning based supervised methods. 

The contribution of this paper is twofold. First, it presents \SentApp, a practical weakly-supervised system that does not require labeled data for training, hence can be rapidly and cost-effectively used across different domains for producing fine-grained sentiment reports. 
Second, it introduces a workflow that enables users to weakly-supervise the system, thus enhancing its precision. This workflow enables users to select an unlabeled input dataset from a completely new domain, produce domain-specific aspect and opinion lexicons, and edit the lexicons. The user can then select an unlabeled target dataset from the same domain, classify it, and obtain a detailed report regarding the positive and negative sentiments expressed towards each aspect in the corpus and browse through the results.

Our system has been successfully deployed in several real-life use cases. One of them is the analysis of social media opinions towards specific aspects of movies, preformed in collaboration with a major entertainment content provider.
Another use case involves measuring the impact of social events, like conventions and conferences, based on opinions published in social media posts.

The system is distributed as open source software under the Apache license as part of NLP Architect by Intel AI Lab.\footnote{\url{http://nlp_architect.nervanasys.com/absa.html}}

\section{System Workflow}
\label{sec:workflow}

\begin{figure*}[t!]
\includegraphics[width=\textwidth]{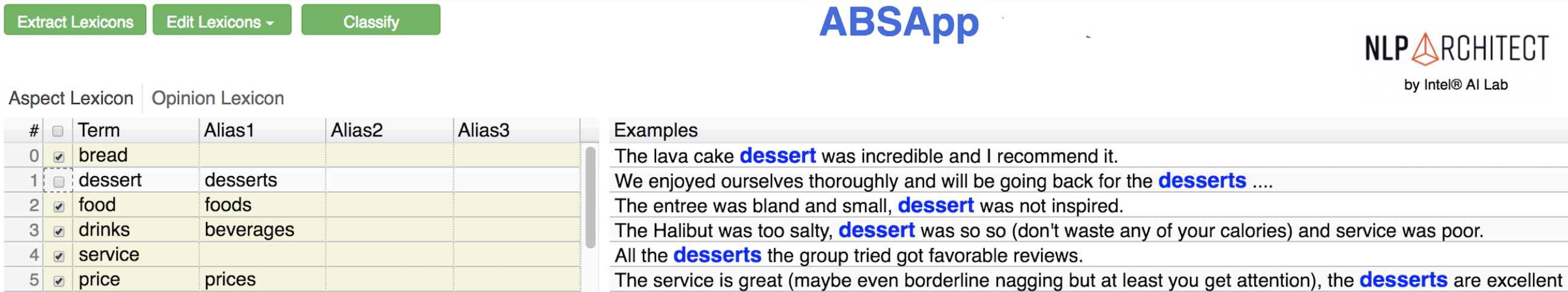}
\centering
\caption{\SentApp user interface for aspect and opinion lexicons management. The figure displays an aspect lexicon related to the restaurants domain.}
\label{UI_1_fig}
\end{figure*}

This section describes the workflow of \SentApp, as depicted in Figure~\ref{workflow_fig}.

\paragraph{Steps 1 \& 2: Selecting a Dataset and Extracting Lexicons.}
The first step of the flow is to select an input dataset for lexicon extraction, performed by clicking the {\it Extract lexicons} button shown in Figure~\ref{UI_1_fig}. Once a dataset is selected, the system performs the lexicon extraction process. This step extracts aspect terms and produces an aspect lexicon. In addition, this step extracts candidate opinion terms, filters them and estimates their polarity, producing an opinion lexicon (see Section \ref{sec:system_lex_extraction}).  

\paragraph{Step 3: Lexicon Editing.}
\label{sec:LexiconsEditing}
Figure~\ref{UI_1_fig} shows the aspect and opinion lexicon management (editing) user interface. The user can choose to edit an aspect lexicon or an opinion lexicon that was generated in step 2. 
As shown in Figure~\ref{UI_1_fig}, in which the {\it Aspect Lexicon} has been selected, the {\it Term} column displays the aspect terms while the {\it Alias1-3} columns display aspect terms that have the same semantic meaning. 

Upon selecting a specific aspect, the {\it Examples} view on the right-hand side of Figure~\ref{UI_1_fig} displays text snippets from the input dataset that include this term (highlighted in blue). The {\it Examples} view enables the user to verify that the selected term indeed functions as an aspect term in various contexts in the domain. Based on this, the user can delete (by unchecking the term's checkbox), add or modify the lexicon items. 
The recommended best practice is to keep relevant aspect terms and delete non-relevant aspect terms. For example, keep terms like 'service' and 'decor' and delete terms like 'time' and 'city' from an aspect lexicon related to restaurant reviews.  

In addition, the user can group together synonym aspects like 'drinks' and 'beverages' (see Figure~\ref{UI_1_fig}). Finally, the user can save the edited lexicon.
The opinion lexicon editor (not shown) functions similarly to the aspect lexicon editor except that it includes a {\it Polarity} column and a {\it Score} column (see Section \ref{sec:system_lex_extraction}) instead of the {\it Alias} columns. Both the polarity and the score can be edited by the user.

\begin{figure*}[h!]
\includegraphics[width=\textwidth]{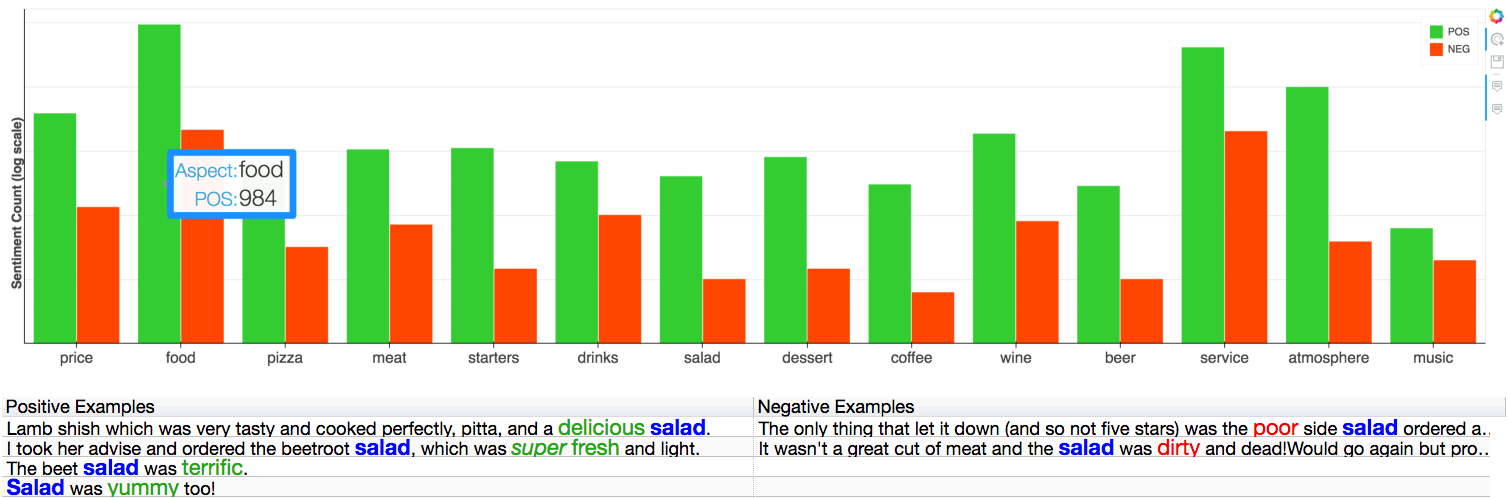}
\centering
\caption{\SentApp user interface for displaying the accumulated amount of positive \& negative sentiment per aspect (top) and sentences containing sentiment towards a specific aspect (bottom).\newline}
\label{UI_2_fig}

\end{figure*}

\paragraph{Steps 4 \& 5: Selecting a Target Dataset and Performing Sentiment Classification.}
A target dataset and its classification are performed by clicking the {\it Classify} button in Figure~\ref{UI_1_fig}. Once the dataset is selected the system starts the sentiment classification process (see Section \ref{sec:system_classification}).

\paragraph{Step 6: Results Visualization.}
Figure~\ref{UI_2_fig} shows the output of the sentiment classification process. The upper part of the figure displays the count of positive and negative sentiment mentions detected in the target dataset towards each aspect, as green and red bars.
Hovering over a green(red) bar displays the count of the positive(negative) sentiment mentions towards a specific aspect (see the blue rectangle in Figure~\ref{UI_2_fig}). The displayed value is an aggregation of the sentiment mention count towards the aspect term and towards all of its {\it Alias} terms (as in Figure~\ref{UI_1_fig}).

Upon clicking a bar related to a specific aspect, a list of sentences containing positive and negative sentiment towards that aspect is displayed with the aspect terms colored blue and the positive and negative opinion terms colored green and red, respectively (lower part of Figure~\ref{UI_2_fig}). This view enables the user to drill-down into the results and extract further insight.

\section{Algorithmic Components}
\label{sec:algo}
Our algorithmic approach is based on using unlabeled data from a new target domain to co-extract aspect and opinion terms, in order to generate domain-specific aspect and opinion lexicons (Section \ref{sec:system_lex_extraction}). Those lexicons are then used to extract aspect-opinion sentiment mentions in datasets from the same domain (Section \ref{sec:system_classification}).

\subsection{Lexicon Extraction} 
\label{sec:system_lex_extraction}
\paragraph{Pre-processing.} The first lexicon extraction step includes applying tokenization, part-of-speech tagging \footnote{Performed by spaCy (\url{https://spacy.io/}).} and dependency parsing to the input data. For dependency parsing, we used the Bi-LSTM parser proposed by~\citet{BIST_Parser}.

\paragraph{Aspect and Opinion Term Extraction.} This step is based on applying the bootstrap opinion and aspect term co-extraction using the dependency relation rules algorithm, proposed by~\citet{Qiu_and_Lui2011double_prop}.
The bootstrap process is initialized with a seed lexicon of generic opinion terms. New aspect and opinion terms are extracted based on the seed lexicon and the dependency relation rules. The extracted terms are then added to the seed lexicon, and used for extracting additional terms in the next iteration. 
In order to initialize the bootstrap process, we used the opinion lexicon generated by~\citet{Hu_and_Lui2004rule_based}, which contains around 6800 opinion terms along with their sentiment polarity.
Table \ref{table:acq_rules} shows two of the 8 rules that are used along with example sentences. The example for rule {\it R1} illustrates the extraction of the aspect term {\it decor} based on the known opinion term {\it nice}, while the example for rule {\it R2} illustrates the extraction of the opinion term {\it tasty} based on the known aspect term {\it food}. 

\addtolength{\parskip}{-0.5mm}
\begin{table}[H]
   \small 
   \centering 
   \begin{tabular}{lccr} 
   \toprule[\heavyrulewidth]\toprule[\heavyrulewidth]
   \textbf{RuleID} & \textbf{Rule} & \textbf{Example} \\ 
   \midrule
   R1 & O $\xrightarrow{amod}$ A(NN) & nice decor \\     &        &  (nice $\xrightarrow{amod}$ decor)  \\
   R2 & A $\xrightarrow{nsubj}$ O(ADJ) & the food was super tasty \\
   &        &  (food $\xrightarrow{nsubj}$ tasty)  \\
   \bottomrule[\heavyrulewidth] 
   \end{tabular}
   \caption{Examples of the opinion and aspect terms extraction rules. {\it O} represents an opinion term and {\it A} represents an aspect term.} 
   \label{table:acq_rules}
\end{table}

\paragraph{Opinion Lexicon Scoring and Filtering.} This step aims to filter the noisy candidate opinion terms extracted by the bootstrap process. It is based on using an MLP classifier for generating a score that represents the probability that a candidate is indeed an opinion term. Candidate opinion terms are qualified as opinion terms if their classification score exceeds a threshold~\footnote{The threshold's value was empirically determined based on precision-recall tradeoff.}. 

The MLP classifier input features consists of the candidate term word embedding\footnote{\label{note1}We used Stanford Glove embeddings \url{https://nlp.stanford.edu/projects/glove/}} and the mean, standard-deviation, maximum and minimum word-embedding cosine similarities between the candidate term and a pre-determined set of generic opinion terms. 
The MLP consists of a single hidden layer and is trained once for a binary classification task using manually labeled data that consists of a set of opinion terms (positive class) and a set of non-opinion terms (negative class) from a specific domain\footnote{This training data can be downloaded from \url{https://github.com/NervanaSystems/nlp-architect/blob/master/nlp_architect/models/absa/train/lexicons/RerankTrainingData.csv}}. Once the model is generated it is then used for grading candidate opinion terms extracted in other domains. It is reasonable to use such model across domains, since the classification features represent semantic similarity levels that are robust across domains. 

\paragraph{Opinion Polarity Estimation} 
The goal of this step is to set the binary sentiment polarity (positive or negative) of the opinion terms. Following \citet{Polarity_Estimation}, an opinion term polarity is assigned based on estimating whether it is semantically closer to a set of generic positive opinion terms or to a set of generic negative opinion terms. 
To produce those sets we used a subset of 47 positive terms and a subset of 47 negative terms derived from the opinion lexicon generated by~\citet{Hu_and_Lui2004rule_based}. 
The semantic similarity between an opinion term and a positive or negative set is estimated by averaging the cosine similarity of the embedding of the opinion term and the embedding of each one of the terms in the positive or negative set. 

In this module we used pre-trained embeddings\footref{note1} which produce overall good results but raise a drawback; 
some opinion terms may convey different sentiment polarities in different domains (e.g. 'delicate movie' (positive) vs. 'delicate cellphone' (negative)), while a pre-trained embeddings setup is only capable of setting a single polarity per opinion term. A suggested solution is to adapt the embeddings to the target domain or to use context embeddings. We intend to address this challenge in future work.

\subsection{Sentiment Classification}
\label{sec:system_classification}
Sentiment classification uses the opinion and aspect lexicons for detecting aspect-opinion pairs (sentiment mentions) within the input target dataset, and determining their sentiment polarity. 
Aspect-opinion pairs are extracted based on detecting a direct or second-order dependency relation of any type, between them. The aspect-opinion pair polarity is assigned according to the polarity of the opinion term. 
This step also uses a pre-determined negation lexicon containing negation terms. Upon detecting a direct dependency relation between a negation term and the opinion term, the aspect-opinion pair polarity is reversed.  

\section{Evaluation}
\label{sec:evaluation}
Our evaluation objective is to show that the different algorithmic steps, namely, lexicon extraction and sentiment classification, produce usable results. An additional objective is to show that weak supervision of an aspect lexicon that was generated in an unsupervised manner produces comparable results to the recent transfer-learning based methods \cite{Ding2017crossdomain, Wang2018cross_domain}. 
For this purpose, we leveraged the data of SemEval 2014 task 4 ~\cite{pontiki_etal_2014_semeval}, which tests the two main ABSA sub-tasks: aspect term extraction and aspect term polarity detection.

\paragraph{Datasets.} 
The performance of the algorithm was evaluated using data from two different domains: restaurant reviews and laptop reviews. Those two domains are disjoint and therefore demonstrate the robustness of our system. Following previous work, the restaurant reviews dataset consists of the restaurant reviews from SemEval 2014 task 4 subtask 1~\cite{pontiki_etal_2014_semeval} and from SemEval 2015 task 12 subtask 1~\cite{pontiki_etal_2015_semeval}. The laptop domain consists of the laptop reviews from SemEval 2014 task 4 subtask 1. The gold data includes manual labeling of the spans of aspect term mentions within each sentence in the dataset as well as the sentiment polarity (positive, negative, conflict or neutral) related to each aspect. The two domains consist of a total of 5841 and 3614 sentences, respectively.

\paragraph{Experimental Setup.}  
Following the first two subtasks of the SemEval 2014 task 4, our experiment is split into two parts: First, we evaluate aspect term extraction by generating an aspect lexicon, using it for detecting aspect terms within the test set and comparing it against the gold labels. Second, we evaluate the sentiment polarity detected towards each extracted aspect by comparing between the aspect-opinion pairs detected with their assigned polarity and the gold labels.

The data from each domain was randomly split into 75\% training and 25\% testing. The training data was used (ignoring its annotation) for generating the domain-specific opinion and aspect lexicons according to workflow steps 1-3 of Figure~\ref{workflow_fig}. As a baseline to the aspect term extraction evaluation, we tested the unsupervised output of \SentApp, in which the lexicons were not manually edited. We also tested the weakly-supervised output of the system, in which the aspect lexicon was edited ; this manual process, which took about 15 minutes, involved deleting aspects that are non-relevant to the domain
(see step 3 in Section \ref{sec:LexiconsEditing} for detailed description of this process).

Following the settings in prior aspect term extraction work, only exact matches between the predicted aspects and gold labels are counted as correct. We also added a more lenient metric where partial matches are counted as correct, since, for many practical usages, partial matches are sufficient for extracting valuable insight. For example, in the restaurant domain, 'service' and 'waiting service' can be counted as the same aspect. This lenient metric was also used for the aspect term polarity evaluation task.

\paragraph{Results.} 
Table~\ref{table:ATE_table} shows an F1 score evaluation of the aspect term extraction task. It includes a comparison between
the unsupervised output of the system (`ABSApp-unsup'), its weakly-supervised output (`ABSApp-wksup') and two transfer-learning based methods: `Hier-joint' by \citet{Ding2017crossdomain} and `RNSCN' by \citet{Wang2018cross_domain}. `\SentApp-wksup Ln' represents the weakly-supervised system lenient matches. 

It is noticeable from Table~\ref{table:ATE_table} that the results of the unsupervised output of the system (`ABSApp-unsup') are noisy, but that the weakly-supervised output results (`ABSApp-wksup') are quite comparable to the cited transfer-learning based methods, however, the latter require annotated data from a source domain (the results shown are averaged across tests using data from 2 different annotated source domains), whereas \SentApp relies on a short weak supervision process but does not require any labeled data, which is often unavailable in applied industrial settings.

Table~\ref{table:ATP_table} shows an evaluation of the weakly-supervised \SentApp lenient performance of the aspect term polarity task. This task relates to the sentiment polarity (positive, negative) detected towards each extracted aspect mention, hence it reflects the quality of both algorithmic components: aspect and opinion lexicon extraction (Section \ref{sec:system_lex_extraction}) and sentiment classification (Section \ref{sec:system_classification}).

It is seen from Table~\ref{table:ATP_table} that although the recall in both tests is not high (because it reflects correct detection of an aspect term, an opinion term and a relation between them), the precision is above 70\%. Such precision is often sufficient for practical purposes. 
Note that there is no transfer-learning work related to the aspect term polarity task, therefore no benchmarks to other systems are shown.

\vspace*{-5pt}
\begin{table}[h]
   \small 
   \centering 
   \begin{tabular}{lccr} 
   \toprule[\heavyrulewidth]\toprule[\heavyrulewidth]
   \textbf{Model} & \textbf{Restaurants} & \textbf{Laptops}  \\ 
   \midrule
   
   Hier-Joint\tnote{\textdagger} & 48.0 & 34.2 \\
   RNSCN\tnote{\textdagger}  & 51.5 & 45.9      \\ \midrule
   \SentApp-unsup & 43.5 & 23.2 \\
   \SentApp-wksup & 51.1 & 40.1   \\ \midrule \midrule
   \SentApp-wksup Ln & 66.9 & 58.2\\
   \bottomrule[\heavyrulewidth] 
   \end{tabular}
   \caption{Aspect term extraction evaluation (F1 score). 
    \tnote{\textdagger} average performance over evaluations using different random dataset splits of the test data, as reported by \citet{Wang2018cross_domain}.}
   \label{table:ATE_table}
\end{table}

\vspace*{-15pt}
\begin{table}[h]
   \small 
   \centering 
   \begin{tabular}{lccr} 
   \toprule[\heavyrulewidth]\toprule[\heavyrulewidth]
   \textbf{Domain} & \textbf{Precision} & \textbf{Recall} & \textbf{F1 score} \\ 
   \midrule
   Restaurants & 70.3 & 44.5 & 54.6 \\
   Laptops & 72.7 & 27.6 & 40.1     \\
   \bottomrule[\heavyrulewidth] 
   \end{tabular}
   \caption{Weakly-supervised ABSApp aspect term polarity evaluation.} 
   \label{table:ATP_table}
\end{table}

\section{Field Use Cases}\label{sec:usecases}
This section describes two use cases in which \SentApp has been successfully used. 

\paragraph{Movie Reviews Analysis.}
One of the main predictors for the commercial success of a movie is the estimated hype effect of the movie's pre-release in social media as measured using sentiment analysis~\cite{Movie_success_prediction}. 
\SentApp was used in collaboration with a major entertainment content provider for analyzing audience opinion in social media towards movies and trailers. The system detected the different characters, actors, scenes and music as aspects, and produced fine-grained sentiment reports periodically. These reports were used to fine-tune the content of future movie trailer releases.

\paragraph{Convention Impact Analysis.}
Analysis of sentiment towards different aspects is useful also for measuring the impact of professional events, determining user impressions, and acting accordingly. \SentApp was used during the 2018 Intel AI development convention in San Francisco~\footnote{\url{https://newsroom.intel.com/press-kits/2018-ai-devcon/\#gs.bejr0q}} to extract aspects related to the convention and to analyze the sentiment towards them based on Twitter feeds. 
The system detected aspects like sessions, keynotes, demos, venue, etc., and provided the event organizers with valuable information regarding the level of positive/negative sentiment towards them. In addition, it enabled organizers to browse through sentences containing sentiment towards specific aspects and draw conclusions as to what should be changed and what should be continued at current and future conventions.
 
\section{Conclusion}
\label{sec:conclusion}
We presented \SentApp, a weakly-supervised system for Aspect Based Sentiment Analysis. We showed that weak supervision of lexicons, which were generated in an unsupervised manner, produces comparable results to recent supervised transfer-learning based methods. This enables, rapid and cost-effective use across different domains in applied setups where labeled data is often unavailable.

\bibliography{references.bib}
\bibliographystyle{acl_natbib}

\end{document}